\documentclass{article}

\usepackage{arxiv}

\usepackage[utf8]{inputenc} 
\usepackage[T1]{fontenc}    
\usepackage{hyperref}       
\usepackage{url}            
\usepackage{booktabs}       
\usepackage{amsfonts}       
\usepackage{nicefrac}       
\usepackage{microtype}      
\usepackage{lipsum}		
\usepackage{graphicx}
\usepackage{natbib}
\usepackage{doi}

\usepackage{algorithm}
\usepackage{algpseudocode}
\usepackage{amsmath}

\title{Transformer Layer Injection: A Novel Approach for Efficient Upscaling of Large Language Models}

\author{ \href{https://orcid.org/0000-0002-4363-2177}{\includegraphics[scale=0.06]{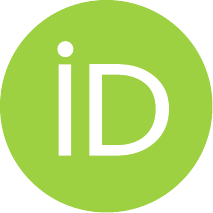}\hspace{1mm}James Vo}\thanks{Anh-Dung Vo} \\
	DocumentAI Team\\
	AGILESODA INC.\\
	Seoul, 06149, South Korea \\
	\texttt{anhdungitvn@agilesoda.ai} \\
}

\renewcommand{\shorttitle}{Transformer Layer Injection (TLI)}

\hypersetup{
pdftitle={Transformer Layer Injection: A Novel Approach for Efficient Upscaling of Large Language Models},
pdfsubject={q-bio.NC, q-bio.QM},
pdfauthor={David S.~Hippocampus, Elias D.~Striatum},
pdfkeywords={First keyword, Second keyword, More},
}

\begin{document}
\maketitle

\begin{abstract}

In this paper, we propose Transformer Layer Injection (TLI), a novel method for efficiently upscaling large language models (LLMs) while minimizing computational costs and maintaining model performance. Model scale is a key factor in enhancing the quality of machine learning models, and TLI addresses the challenge of scaling by reducing initial loss, minimizing fine-tuning requirements, and preserving model complexity. Our approach improves upon the conventional Depth Up-Scaling (DUS) technique by injecting new layers into every set of K layers, enabling hidden representations to pass through transformer blocks with minimal disruption. We compare TLI with existing approaches, including Mixture of Experts (MoE) and DUS, and validate its efficiency through experiments on small LLMs (LLama3 1B, 3B, and 8B). Results show that TLI achieves better initialization, requires fewer training steps, and delivers superior accuracy on tasks such as KoBEST and KMCQA, with models performing effectively even without additional training. TLI is demonstrated to be both data-efficient and cost-effective, significantly outperforming existing methods. Its scalability and simplicity make it a promising solution for upscaling transformer-based models, with potential applications in scaling models from 10B to 405B parameters.

\end{abstract}


\section{Introduction}

In recent years, large language models (LLMs) have become foundational to advancements in natural language processing (NLP), powering applications such as machine translation, question answering, and text generation. These models, exemplified by architectures like GPT (\cite{openai2024gpt4technicalreport}), LLama (\cite{dubey2024llama3herdmodels}), and Mistral (\cite{jiang2023mistral7b}), have demonstrated that scaling up model size directly correlates with improvements in performance across a wide range of tasks (\cite{kim2024solar107bscalinglarge, komatsuzaki2023sparseupcyclingtrainingmixtureofexperts, tan2020efficientnetrethinkingmodelscaling, fedus2022switchtransformersscalingtrillion}). The growing consensus is that larger models, when trained on vast amounts of data, exhibit better generalization, understanding, and generation capabilities. However, this trend towards larger models introduces significant computational and architectural challenges, particularly when attempting to upscale models in a resource-efficient manner.

Upscaling a model's parameters from x billion to y billion, where y exceeds x, without incurring substantial computational overhead presents a central challenge. Traditional approaches to model scaling, such as increasing depth or width, often result in inefficient initializations, high memory demands, and extended fine-tuning processes. These challenges are compounded by the necessity to maintain model complexity while minimizing training costs and time, factors that become especially critical when scaling models beyond tens of billions of parameters. Existing methods, such as Mixture of Experts (MoE) (\cite{fedus2022switchtransformersscalingtrillion}) and Depth Up-Scaling (DUS) (\cite{kim2024solar107bscalinglarge}), have been proposed to address these issues, each offering advantages and limitations. MoE models, for instance, enable sparsity in layer utilization, improving computational efficiency but at the cost of increased architectural complexity and fine-tuning requirements. DUS, on the other hand, is simple to implement but leads to suboptimal initialization due to mismatches in layer characteristics, resulting in increased training loss.

In this context, we introduce Transformer Layer Injection (TLI), a novel approach aimed at efficiently upscaling LLMs while preserving model integrity and minimizing both computational and training demands. The TLI method enhances traditional upscaling techniques by injecting layers within predefined intervals of existing layers, ensuring that the hidden representations within the transformer blocks pass with minimal disruption. This approach provides better model initialization, reducing initial loss and allowing for faster convergence during training. Unlike other methods, TLI offers a solution that is both scalable and practical, delivering superior performance without the need for additional fine-tuning or increased architectural complexity.

This paper provides a detailed exploration of the TLI method and its application to LLM upscaling. We conduct experiments on smaller-scale LLMs, including LLama3 models with 1B, 3B, and 8B parameters, comparing TLI with MoE and DUS across several benchmarks. The results demonstrate that TLI outperforms both methods in terms of initialization efficiency, data utilization, and model accuracy, with a significant reduction in the number of training steps required to reach optimal performance. Furthermore, TLI proves to be a cost-effective and scalable solution, capable of supporting upscaling efforts for models as large as 405B parameters.

The remainder of this paper is structured as follows: Section 2 reviews the current state-of-the-art methods for model upscaling, highlighting their strengths and weaknesses. Section 3 introduces the TLI methodology, detailing its implementation and theoretical foundations. In Section 4, we present the experimental setup, followed by results and analysis in Section 5. Finally, Section 6 concludes with a discussion of the broader implications of TLI and outlines directions for future research.

\section{Related Work}
\label{sec:related_work}

The rapid development of large language models (LLMs) has introduced significant advancements in natural language processing (NLP) tasks, with models like GPT, LLama, and Mistral demonstrating state-of-the-art performance across various domains. The trend toward larger models, with parameter counts extending into the hundreds of billions, has proven effective for improving generalization and task-specific performance. However, this scaling comes with significant computational and architectural challenges. As such, various techniques have been proposed to address the efficiency issues associated with scaling LLMs, including Mixture of Experts (MoE) and depth-wise scaling strategies. This section reviews the key methods that have informed the development of Transformer Layer Injection (TLI).

One prominent approach is the Depth Up-Scaling (DUS) method introduced by Kim et al. in the SOLAR 10.7B model, which focuses on increasing the depth of transformers by duplicating top and bottom layers, effectively expanding the model’s depth while maintaining simplicity. DUS demonstrates that scaling model depth without significantly altering model architecture can lead to improved performance in LLMs like SOLAR 10.7B, which outperformed prior models on a wide range of NLP tasks (\cite{kim2024solar107bscalinglarge}). Despite its simplicity, however, DUS suffers from suboptimal initialization, as duplicating layers with varying characteristics introduces training inefficiencies, often resulting in increased initial loss. This limitation is one of the key challenges that TLI seeks to address, as TLI’s method of injecting new layers at regular intervals mitigates the initialization issues caused by abrupt layer duplication in DUS.

Another influential approach is the Mixture of Experts (MoE) model, which has gained traction as an efficient strategy for managing computational costs in LLM scaling. MoE introduces sparse model activation by routing tokens to specific “experts,” where each expert consists of a subset of the model's layers (\cite{fedus2022switchtransformersscalingtrillion}). This selective activation allows MoE models, such as the Switch Transformer, to manage an enormous number of parameters without a proportional increase in computational cost. While MoE has demonstrated significant improvements in pre-training speed and resource efficiency, especially in large-scale multilingual settings, it introduces architectural complexity that makes fine-tuning and inference more difficult (\cite{fedus2022switchtransformersscalingtrillion}). The routing of tokens to experts can also result in instability during training, requiring careful tuning and additional communication overhead. In contrast, TLI avoids the architectural overhead of MoE while still providing a scalable solution to efficiently increase model size.

The concept of sparsity is further explored in Sparse Upcycling, where pre-trained dense models are converted into sparse MoE models, allowing for more efficient computation without training from scratch. Sparse Upcycling reduces the cost of training large models by leveraging already trained dense checkpoints and converting them into sparsely activated experts (\cite{komatsuzaki2023sparseupcyclingtrainingmixtureofexperts}). This method has proven effective in multiple domains, including both natural language and vision tasks, yielding significant efficiency improvements. However, like other sparse approaches, it requires complex modifications to the original dense models and involves intricate model management during the training and inference phases. TLI, in comparison, takes a different route by preserving the dense architecture while still offering an efficient mechanism for scaling LLMs through layer injection.

In addition to these LLM-specific strategies, scaling methods developed for **Convolutional Neural Networks (CNNs)** provide useful insights into efficient model scaling. **EfficientNet**, a pioneering scaling strategy for CNNs, introduced a compound scaling method that simultaneously adjusts the network’s depth, width, and resolution (\cite{tan2020efficientnetrethinkingmodelscaling}). This balanced scaling approach demonstrated that improvements in one dimension often require corresponding adjustments in others to optimize performance. While EfficientNet is designed for CNNs, the principle of balanced scaling is analogous to LLM scaling efforts, where depth and width adjustments must be carefully managed to avoid inefficiencies. TLI’s approach to layer injection takes inspiration from such balanced scaling efforts by ensuring that new layers are introduced systematically across the model to minimize disruption to hidden representations.

Switch Transformers have also significantly advanced the scaling of language models to the trillion-parameter level by leveraging sparse MoE techniques to decouple parameter size from computational cost. Fedus et al. (\cite{fedus2022switchtransformersscalingtrillion}) demonstrated how sparsely activated models could be trained faster and at a lower cost than dense models, achieving up to a 7x increase in pre-training speed. However, as with other MoE approaches, Switch Transformers require sophisticated routing mechanisms that introduce complexity into both the training and inference stages. In contrast, TLI's method offers a simpler alternative by focusing on the injection of layers into dense transformer models, achieving efficiency without increasing architectural complexity.

Overall, these methods represent important advancements in the field of LLM scaling, but each approach comes with trade-offs in terms of complexity, initialization efficiency, and computational demands. TLI builds upon the strengths of these prior methods while addressing their limitations, offering a more straightforward and effective solution for scaling LLMs. By systematically injecting layers within transformer blocks, TLI ensures better initialization, reduces training steps, and provides a scalable framework for expanding LLMs to hundreds of billions of parameters without the need for extensive fine-tuning or increased architectural complexity.

\section{Methodology}
\label{sec:methodology}

In this section, we describe the Transformer Layer Injection (TLI) method, detailing its core motivation, problem statement, and the implementation steps involved in achieving efficient model upscaling. TLI is designed to mitigate the challenges associated with traditional upscaling approaches, such as Depth Up-Scaling (DUS) and Mixture of Experts (MoE), by ensuring better initialization, minimal computational overhead, and effective training performance.

\subsection{Motivation}
\label{subsec:motivation}
Scaling up large language models (LLMs) is essential for improving performance in various natural language processing tasks. However, the conventional scaling techniques often require extensive computational resources and time. The motivation behind TLI is to introduce a method that allows for efficient model expansion by systematically injecting layers into the transformer architecture. By strategically inserting new layers at predefined intervals, TLI minimizes disruption to hidden representations, resulting in better model initialization and reduced training requirements.

\subsection{Problem Statement}
\label{subsec:problem_statement}
The challenge lies in scaling a given model from x billion parameters to y billion parameters, where y exceeds x, in a resource-efficient manner. This requires maintaining model performance, minimizing computational costs, and avoiding extensive fine-tuning. Traditional approaches like DUS result in suboptimal initialization and increased training loss, while MoE introduces architectural complexity. The objective of TLI is to provide a scalable, efficient method to upscale LLMs without these drawbacks.

\subsection{Definition of Efficiency}
\label{subsec:definition_of_efficiency}
Efficiency in the context of LLM upscaling is defined by the following criteria:

1. \textbf{Minimal computational overhead}: The method should not significantly increase memory or computational requirements.

2. \textbf{Reduced initial loss}: The new layers should integrate smoothly with the existing layers, ensuring minimal loss during initialization.

3. \textbf{Limited fine-tuning requirements}: The model should achieve high performance with minimal additional training.

4. \textbf{Preservation of model complexity}: TLI aims to maintain the original architecture's simplicity while improving scalability and performance.

\subsection{Transformer Layer Injection (TLI) Approach}
\label{subsec:transformer_layer_injection_tli_approach}

\subsubsection{Layer Injection Strategy}
The central idea of TLI is to inject new layers within a set of existing layers. For every K layers in the original model, a new layer is introduced by duplicating the previous transformer layer. This approach allows hidden representations to pass through the model with minimal disruption, enabling a smoother transition between layers and avoiding the initialization problems seen in DUS.

The process involves the following steps:

1. \textbf{Identifying layer injection points}: The model is divided into intervals, where every K layers a new layer is injected.

2. \textbf{Duplicating transformer layers}: For each injection point, the previous transformer layer is duplicated, preserving the model's hidden representations.

3. \textbf{Ensuring model stability}: The new layers are initialized to ensure that the model remains stable during training and requires fewer fine-tuning steps.

\subsubsection{How It Works}

\begin{figure}[ht]
\centering
\includegraphics[width=0.80\linewidth]{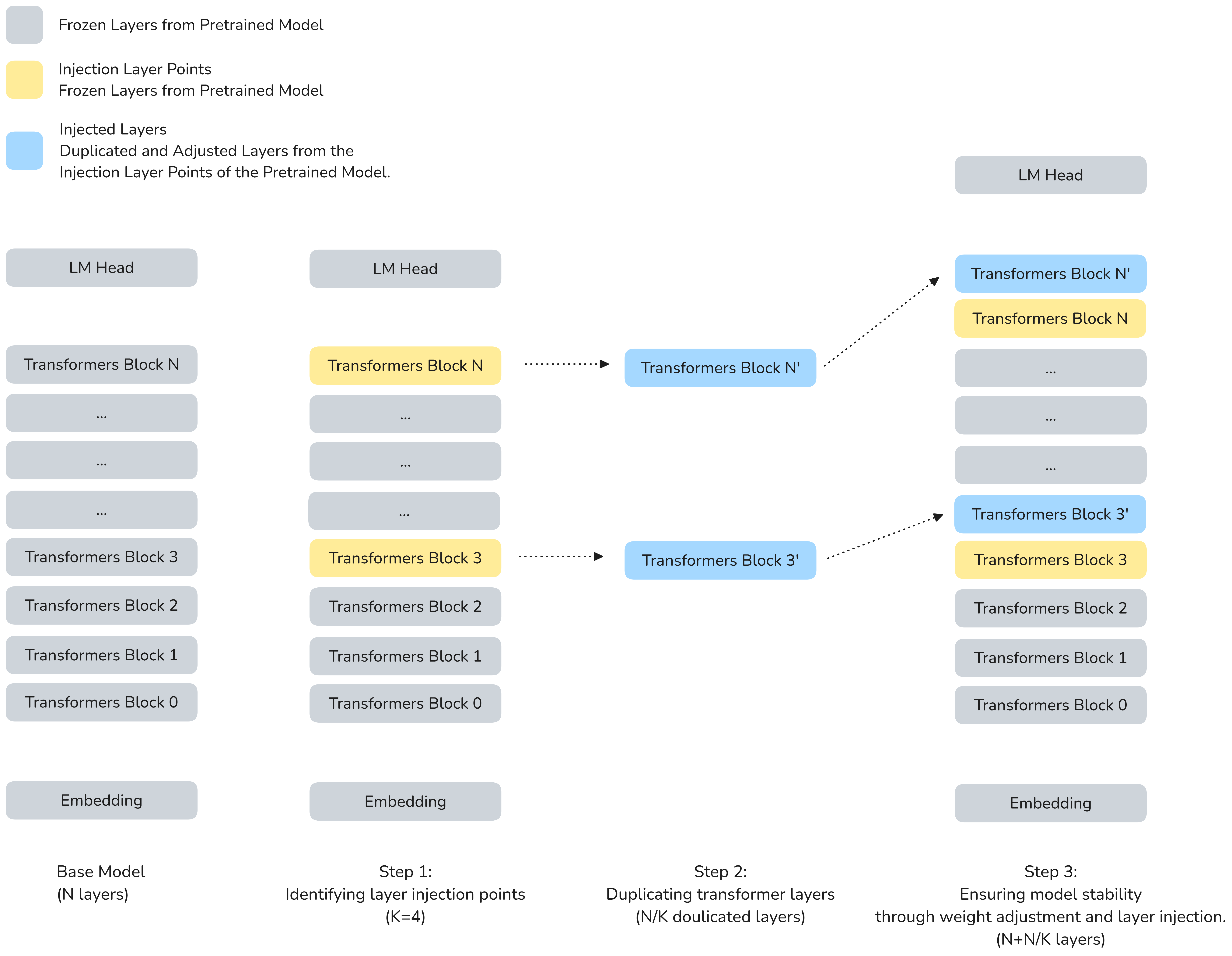}
\caption{Proposed Transformer Layer Injection (TLI) Method}
\label{fig:fig_proposed_tli}
\end{figure}

The TLI is illustrated in Figure~\ref{fig:fig_proposed_tli} and formally presented in Algorithm~\ref{algorithm:TransformerLayerInjection}, which outlines its key steps.

\begin{algorithm}
\caption{Transformer Layer Injection (TLI)}
\label{algorithm:TransformerLayerInjection}
\begin{algorithmic}[1]
\State \textbf{Input:} Model $M$, original number of layers $L_{orig}$, desired number of layers $L_{new}$, checkpoint $ckpt$
\State Initialize layer injection interval: $split = \frac{L_{orig}}{L_{new} - L_{orig}}$
\State Initialize $requires\_grad\_layers = []$, $layer\_cnt = 0$
\State Initialize empty dictionary $output$ to store modified state dict
\For{each layer $i = 0, 1, ..., L_{orig}-1$}
    \For{each key $k$ in $ckpt$}
        \If{$k$ contains $layers.i$}
            \State Replace $i$ with $layer\_cnt$ in $k$ and store in $output$
        \EndIf
    \EndFor
    \State Increment $layer\_cnt$
    \If{$(i + 1) \% split == 0$}
        \For{each key $k$ in $ckpt$}
            \If{$k$ contains $layers.i$}
                \If{$k$ contains 'down\_proj' or 'o\_proj'}
                    \State Insert a zero tensor for the new layer in $output$
                \Else
                    \State Copy the original layer's parameters to the new layer in $output$
                \EndIf
            \EndIf
        \EndFor
        \State Append $layer\_cnt$ to $requires\_grad\_layers$
        \State Increment $layer\_cnt$
    \EndIf
\EndFor
\State Verify $layer\_cnt == L_{new}$
\For{each key $k$ in $ckpt$ not containing 'layers'}
    \State Copy the key-value pair to $output$
\EndFor
\State Save the modified model to $output\_dir$
\State Update model config: set new layer count, $requires\_grad\_layers$, and other parameters
\State Save the updated model configuration
\end{algorithmic}
\end{algorithm}

\subsubsection{Training-wise Strategy}
It is important to note that the output model generated by the TLI method is capable of functioning effectively without the need for additional training. However, in scenarios where further training is required, such as 1) enhancing model performance for a specific task, or 2) incorporating new knowledge related to a novel task by adjusting the newly introduced N/K layers, we propose a cost-effective, training-wise strategy that facilitates rapid optimization of the model.

As depicted in Figure 1, the TLI method employs a transformer-based model with N transformer layers as input and produces an augmented model comprising N + N/K layers, where K is a parameter that defines the Injection Layer Points. Crucially, within the N + N/K layers, only N/K layers are subject to new weight initialization, while the majority of the N layers are transferred from the original pretrained model.

1. \textbf{Step 1: Training of the new N/K layers}: The first step involves freezing the entire model, with the exception of the newly injected N/K layers. These newly introduced layers are then trained to ensure efficient adaptation to the specific requirements of the task.

1. \textbf{Step 2: Training the entire model}: In the second step, the entire model (N + N/K layers) is fine-tuned using efficient techniques, such as LoRA (\cite{hu2021loralowrankadaptationlarge}) or QLoRA (\cite{dettmers2023qloraefficientfinetuningquantized}), which allow for rapid convergence to an optimal state.

\subsubsection{Computational Advantages}
By injecting new layers at regular intervals rather than duplicating entire sections of the model, TLI reduces the number of additional computations required compared to DUS. Additionally, since the injected layers are derived from the existing transformer architecture, the method avoids the architectural complexities introduced by MoE.

\subsection{Strategy for Experimental Setup}
We evaluate the performance of TLI using LLama3 models of varying sizes, specifically focusing on models with 1B, 3B, and 8B parameters. The experiments were conducted by applying both TLI and DUS methods to scale these models, followed by a comparison of training steps, initialization loss, and overall performance.

Strategy for Scaling the Model

For each model, the following steps were executed:

1. \textbf{Model Initialization}: The base model was loaded, and the original number of layers was identified.

2. \textbf{Layer Injection}: TLI was applied by injecting layers at predefined intervals, resulting in upscaled models with a higher number of parameters.

3. \textbf{Performance Evaluation}: The modified models were fine-tuned using tasks such as KoBEST (\cite{kim2022kobestkoreanbalancedevaluation}) and KMCQA (Korean Multi-Choice Question Answering reference being updated) to assess performance improvements.

\subsection{Strategy for Results and Observations}

1. \textbf{Initialization Efficiency}: TLI demonstrated superior initialization efficiency, with the injected layers smoothly integrated into the model architecture. The initialization loss was significantly lower compared to DUS, allowing for faster convergence during training.

2. \textbf{Reduced Training Steps}: Models upscaled using TLI required fewer training steps to achieve comparable or better performance than those upscaled using DUS. This indicates that TLI provides better data efficiency, reducing the time and computational cost associated with training.

3. \textbf{Performance on Benchmark Tasks}: The models upscaled using TLI consistently outperformed those using DUS on tasks such as KoBEST (\cite{kim2022kobestkoreanbalancedevaluation}) and KMCQA (Korean Multi-Choice Question Answering reference being updated). This was achieved without requiring additional fine-tuning, further demonstrating TLI's efficiency in scaling LLMs.

\section{Experimental Results}
In the next section, we will discuss the experimental results in more detail, providing insights into the advantages of TLI over traditional scaling methods.

\section{Conclusion}
TLI presents a novel, efficient solution for scaling LLMs. By injecting layers at regular intervals, it ensures better model initialization, reduces training steps, and delivers superior performance on key NLP tasks. The simplicity and scalability of TLI make it a promising alternative to existing upscaling methods, offering significant improvements in terms of computational efficiency and model accuracy.

\bibliographystyle{unsrtnat}
\bibliography{references}  






\end{document}